\documentclass[journal]{IEEEtran}
\ifCLASSINFOpdf
\else
\fi
\usepackage{amsmath,amssymb,amsfonts}
\usepackage{algorithmic}
\usepackage{graphicx}
\usepackage{lineno,hyperref}
\usepackage{booktabs}
\usepackage{color}
\usepackage{balance}

\begin{document}
%
\title{Deep Learning Models for Early Detection and Prediction of the spread of Novel Coronavirus (COVID-19)}
%
%
%

\author{Devante~Ayris*, Kye~Horbury*, Blake~Williams*, Mitchell~ Blackney, Celine~Shi~Hui~See*, Maleeha~Imtiaz*, Syed~Afaq~Ali~Shah*$^{+}$~\IEEEmembership{Member,~IEEE}
\thanks{Discipline of Information Technology, Media and Communications,
$^{+}$Harry Butler Institute,
Murdoch University, Australia,               }
\thanks{{*}All the authors have equal contribution.}
}

\maketitle

\begin{abstract}
SARS-CoV2, which causes coronavirus disease (COVID-19) is continuing to spread globally and has become a pandemic. People have lost their lives due to the virus and the lack of counter measures in place. Given the increasing caseload and uncertainty of spread, there is an urgent need to develop machine learning techniques to predict the spread of COVID-19. Prediction of the spread can allow counter measures and actions to be implemented to mitigate the spread of COVID-19. In this paper, we propose a deep learning technique, called Deep Sequential Prediction Model (DSPM) and machine learning based Non-parametric Regression Model (NRM) to predict the spread of COVID-19. Our proposed models were trained and tested on publicly available novel coronavirus 2019 dataset. The proposed models were evaluated by using Mean Absolute Error and compared with baseline method. Our experimental results, both quantitative and qualitative, demonstrate the superior prediction performance of the proposed models. 

\end{abstract}

\begin{IEEEkeywords}
COVID-19 prediction, Machine Learning, Deep Learning, Regression, MAE
\end{IEEEkeywords}

%
\IEEEpeerreviewmaketitle

\section{Introduction}
\label{intro}
COVID-19 is a pandemic that has spread and devastated countries around the world. Even months on from the original outbreak of the virus, it still poses a large threat to everyone around the globe, as with each passing day, the death toll still increases, and more and more cases are identified. Countries have been brought to a standstill as citizens are forced to self-isolate and worldwide economies have come to a halt as a result of the negative impacts on trade and industry.

First discovered in Wuhan City, Hubei Province of China, on the 31st of December 2019, COVID-19 is a respiratory illness with pneumonia-like qualities and was initially thought to be caused by human contact with exotic fauna, eventually resulting in a person-to-person spread. This virus has caused a massive negative international impact and has affected the day-to-day lives of millions of people, through bans on large public gatherings, panic-buying and travel bans.

It is still difficult to predict where and when new cases will appear, and many governments have failed to understand the scale and impact of the virus. The exponential spread of the virus means that until there is a vaccine, or it has been completely removed from the population, it will always pose a threat even in locations with the best circumstances. 

Deep learning has been a growing trend in data analysis and predictive modeling in recent years, and has been termed one of the ten breakthrough technologies \cite{greenspan2016guest}. It is emerging as the leading machine learning tool in computer vision. This data-driven approach has shown a significant improvement in the performance of classification for large scale natural image datasets. For instance, in the 2012 ImageNet LSVRC contest, the first large-scale deep model, achieved considerably lower error rates compared to the previous methods. After that, several deep learning models have been proposed to further decrease their error rate.

Deep Learning has shown unprecedented performance for several computer vision tasks. It learns the most predictive features (learned features) directly from data given a large dataset of labeled examples. In recent years, deep learning techniques have emerged as highly effective methods for prediction and decision-making in a multitude of disciplines including health (hearing aids), computer vision (e.g., object and face identification), \cite{shah2019spatial}, \cite{shah2016iterative}, \cite{shah2017efficient}, \cite{hu20172d}, natural language processing \cite{wang2020object}, \cite{sharif2019lceval}, \cite{sharif2018nneval}, gesture recognition \cite{zhang2018attention}, \cite{zhu2018continuous}, \cite{zhang2017learning}, and robotics \cite{shah2016novel}. 

Inspired by the recent advancement in machine/deep learning, this research hypothesizes that machine learning can be used to predict the spread of the virus and potentially be used to help allocate resources and prepare procedures ahead of time to mitigate the impacts of COVID-19, potentially saving lives.
In this paper, we propose two different techniques to predict the spread of COVID-19. The paper proposes Deep Sequential Prediction Model (DSPM), which benefits from the sequential nature of the data to make accurate prediction about the spread of this disease. The paper also proposes an efficient Non-parametric Regression Model (NRM), which avoids computationally expensive parameter learning process to efficiently predict the spread of COVID-19. The paper also extensively evaluates the proposed models and analyses their viability to predict the spread of COVID-19 through the world’s population. The motivation of this research is to develop artificial intelligence models, which can accurately predict the spread of COVID-19, thus allowing more refined actions and strategies to take place to mitigate, control and contain the virus. 
The contributions of this paper can be summarized as follows:
\begin{itemize}
    \item The paper proposes a deep sequential prediction model (DSPM) to learn distinctive features from the input time series data for accurate prediction of COVID-19 spread
    \item The paper also proposes a non-parametric regression model (NRM) to accurately and efficiently predict the spread of this contagious disease.
    \item Extensive evaluation of the proposed models has been performed on publicly available novel coronavirus dataset. Our experimental results demonstrate the superior performance of the proposed models.
\end{itemize}{}

The rest of this paper is organized as follows. Section 2 presents and discusses the related work. Section 3 presents our proposed techniques to predict the spread of COVID-19. Experimental results are provided in Section 4, which also provides details of the novel Coronavirus dataset. Section 5 provides discussion and analysis about the proposed techniques. The paper is concluded in Section 6.

\section{Literature Review}
\label{sec:1}
In this section, we first present relevant deep learning and machine learning techniques and then discuss the use of machine learning for infectious diseases.  
Machine learning algorithms, which automatically learn features from the input data, have been evolving over many years and continues to provide several benefits in all aspects of global life, such as uses in the medical industry to economics and business. This has led to the creation of different machine learning models to help solve problems, such as to predict the spreading of diseases, in a way that was not possible or was not highly accurate in the past \cite{wu2019machine}. 

Currently, there are several types of machine learning models reported in the literature. For instance, the Decision Tree model is a predictive model, which breaks up and divides the input data and makes decisions based on a given variable. This is continued to be done recursively for each division, creating a tree like structure, until a solution for the original problem is produced \cite{dias2017risk}.  Decision trees have been used successfully for classification and regression tasks in computer vision.

In addition, Neural Networks are also popular machine learning models. Neural networks mimic a nervous system (human brain) \cite{rojas2013neural} \cite{khan2018guide}, where neurons connect to each other to provide an output. A neural network consists of several layers and data is passed through the input layer to the hidden layers and then finally to the output layers. The output neuron with the highest value (also known as weight) decides the actual class/label of the input data \cite{khan2018guide}. 

Support Vector Machine (SVM) is one of the popular machine learning techniques. This model is designed to split the given data into two even classes via the creation of a hyperplane between them. This hyperplane can then be used to predict the trajectory of future pieces of data. SVM, at its core, involves creating iteratively and infinitely increasing multi-dimensional planes until the most optimal separating hyperplane can be placed between two classes of data points as evenly as possible \cite{tang2015inferring}. A hyperplane is placed depending on the support vectors, which are key pieces of data points, that if altered, change the position of the hyperplane. This is found by using the decision boundary, which is the margin (or distance) between the nearest support vectors and the hyperplane. The SVM finds the most optimal decision boundary by the largest marginal distance between the plane and the support vectors. If a hyperplane cannot be found using the dataset on the current plane, then an extra plane is used and the dataset is checked again. For example, the SVM attempts to split the dataset into two dimensions. If no hyperplane is able to be generated,  then the dataset is placed into the three dimensions and checked again. This process continues until a hyperplane is found [13]. This hyperplane can then be used to further show a predicted trend beyond the original data given by following the plane along the axis. 

Bayesian Networks have also received popularity due to their probabilistic predictive nature \cite{friedman2003being}. These machine learning models involve representing a directed acyclic (non-circular) graph structure, where each node of the graph is a decision, a piece of data or an event. A node can be connected to another depending on whether there is a probabilistic dependency between them. This allows for the prediction on whether certain events to occur and the probability between them [5]. 

Long short-term memory (LSTM) neural networks have received attention from the research community because of their capability to process sequential or time series data. LSTM were originally designed to deal with the vanishing gradient problem \cite{hochreiter1997long}. LSTM neural nets are improved recurrent neural networks (RNNs), which allow cells to remember data from the previous cell through the use of a memory gate. A real word analogy of a memory gate is like a solenoid valve in plumbing where the current water pressure dictates how open the valve is, allowing certain water pressure on its output. The memory gate (or forget gate) works by checking to see if the current input data is the same or similar to the current memory input data and adjusts the memory gates data accordingly. The memory gates work with either a sigmoid layer or a sigmoid and tanh layer that outputs a float variable between zero and one, with zero being do not allow old data through and one being let all the data through. Throughout the LSTM cell there are three memory gates that signify the cell's data state of whether it’s new data or old data. Traversing through these three memory gates ends up with the final output for the LSTM cell.

With the rising issue of the Coronavirus infectious disease (and other similar diseases such as SARS and MERS), there have been few studies involving machine learning to predict the recovery of infected patients and study the similarity of SARS virus protein with other viruses. John et al., proposed machine learning techniques to track and analyze the different factors that are involved in the recovery from MERS \cite{john2019main}. SVM, conditional inference tree, naïve Bayes and J48 models were used to determine and predict whether the categories, including gender, age; the patient is a healthcare worker, status at time of identification of disease, the patient had symptoms and whether the patient had any pre-existing diseases or conditions, were an important factor in determining the recovery of a patient from MERS. Their models determined that age, being a healthcare worker, the status at the time of identification and whether they had pre-existing disease are good indicators at predicting the recovery from MERS, with a p-value of 0.001278, 0.001260, 2e-16 and 0.001067, respectively. 

Cai et al., proposed a method to compare the SARS virus proteins to those of other viruses, to predict how many of those proteins are similar with each other \cite{cai2005prediction}. They used an SVM model in conjunction with the sequence comparison method BLAST to predict the functional class of a given protein, such as whether it is a part of the 46 enzyme families, the 21 channel/transporter families or the 5 RNA-binding protein families to name a few. Their evaluation showed that an SVM can accurately predict the functional class of 73\% of known coronavirus proteins.

Tang et al., proposed a machine learning technique to predict the potential animal hosts of the SARS and MERS viruses \cite{tang2015inferring}. Two machine learning models were used, a non-linear SVM using a radial kernel and a Mahalanobis distance (MD) discriminant model, with both using leave-one-out cross-validation of the training data, to determine host candidates. Both models were successful, with the SVM model having a 99.86\% prediction rate in inferring potential hosts, while the MD model having a 98.08\% prediction rate.

In contrast to the existing techniques, this paper proposes deep/machine learning techniques to predict the spread of novel coronavirus COVID-19. The proposed models have been evaluated on 6.4 million confirmed COVID-19 cases. To the best of our knowledge, this is the first research paper reporting machine learning models for the prediction of COVID-19 spread.

\section{Proposed Models}
\label{sec:2}
In this section, we present our proposed prediction models including Deep Sequential Prediction Model (DSPM) and Non-parametric Regression Model (NRM).

\subsection{Deep Sequential Prediction Model (DSPM)}
\begin{figure}
  \includegraphics[width=0.5\textwidth]{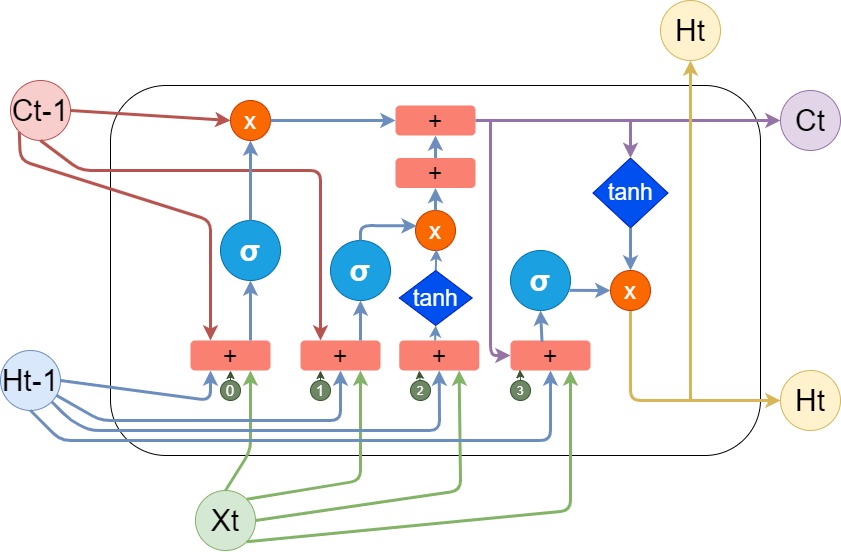}
\caption{Block diagram of the proposed Deep Sequential Prediction Model (DSPM)}
\label{fig:1}       
\end{figure}

Fig. \ref{fig:1} shows the proposed DSPM to predict the spread of COVID-19. As can be noted, our proposed DSPM is a stacked long short-term memory (LSTM) deep neural network. DSPM consists of four stacked LSTMs that feed into each other. These LSTMs contain four hidden layers each (for each stack) that process the data to yield a highly accurate model. We chose stacked LSTMs in our proposed models because the COVID-19 dataset has unknown durations of infection between the countries. This makes training a traditional recurrent neural network (RNN) difficult. This unknown duration period can cause RNN to encounter the vanishing gradient problem, which can completely halt an RNN from further training \cite{pascanu2013difficulty}. On the other hand, a LSTM model is designed to handle this error. In the following, we discuss the different stages of our proposed DSPM. 

\subsubsection{Stage 1}
Given an input data $X_{t}$, this stage (also known as the forget layer) decides whether the cell will throw away the previous data or keep it for modification. It makes this decision through a sigmoid calculation that returns a binary (either one or zero) value. The sigmoid calculation is based on the input vector and the output of the previous block and the memory from the previous block. Therefore, if a new subject is seen, the cell will want to forget the old subject \cite{yan2015understanding}:
\begin{equation}
f_{t}=\sigma\left(W_{f} \cdot\left[H_{t-1}, X_{t}\right]+b_{f}\right)
\end{equation}
where $X_{t}$ is the input vector, $H_{t-1}$ is output of the previous block, $b_{f}$ is a bias term and $\sigma$ is a nonlinear function.

\subsubsection{Stage 2}
The second stage, also known as the input gate layer or new memory valve, processes the data from the previous stage and decides what will be stored in the second memory gate. It is based on a sigmoid layer and a $tanh$ layer. The sigmoid layer works the same way as in Stage 1, while the $tanh$ layer only takes input from the output of the previous block and the input vector. The $tahn$ layer then outputs to the memory gate forming new data  \cite{yan2015understanding}:
\begin{equation}
i_{t} =\sigma\left(W_{i} \cdot\left[H_{t-1}, X_{t}\right]+b_{i}\right) 
\end{equation}
\begin{equation}
\tilde{C}_{t} =\tanh \left(W_{C} \cdot\left[H_{t-1}, X_{t}\right]+b_{C}\right)
\end{equation}

\subsubsection{Stage 3}
In Stage 1, the model decides what data it needs to forget, and in Stage 2 it decides what data it is going to store. With the previous stages deciding what to do with the old data, the model now combines the data to form a new data by combining everything together. To achieve this, it uses the 2 element wise multiplication gates to one summation gate on the memory pipe, as follows: 
\begin{equation}
C_{t}=f_{t} * C_{t-1}+i_{t} * \tilde{C}_{t}
\end{equation}

\subsubsection{Stage 4}
In the final stage, the model finally outputs the data through two channels i.e., the memory channel and the actual output of the cell. First a sigmoid operation is performed that decides about the output. Then the processed memory is put through a $tanh$ nonlinearity. These two operations push through to an element wise multiplication gate. This action is the final output of the cell data. The processed memory then continues onto its own output untouched by this final calculation, while the data output continues after processing  \cite{yan2015understanding}:
\begin{equation}
o_{t}=\sigma\left(W_{o}\left[H_{t-1}, x_{t}\right]+b_{o}\right)
\end{equation}
\begin{equation}
H_{t}=o_{t} * \tanh \left(C_{t}\right)
\end{equation}

\textbf{DSPM Training and Testing}
To train the proposed DSPM, the data that is inputted into the model is first cleaned up from the available time series data. The data is split between country and provinces, and the time series data is then converted to a data frame that includes a date of the confirmed cases. Using empirically selected scalar threshold, this data frame is then converted to 0s and 1s and inputted into the DSPM for its training. DSPM training was found to be faster as the input values are smaller to process. During testing, the model is presented with unseen examples and eventually it outputs its prediction, which are then inverted back to whole numbers via its original scalar threshold.

\subsection{Proposed Non-Parametric Regression Model (NRM)}
In this section, we discuss our proposed non-parametric regression model (NRM). The NRM is based on an additive regression time-series algorithm and uses a decomposable time series model with three major components i.e.,
\begin{equation}
y(t)=g(t)+s(t)+h(t)+\epsilon_{t}
\end{equation}

where $g(t)$ is either linear or a logistic growth curve trend, $s(t)$ are periodic changes, $h(t)$ captures irregular effects, and $\epsilon_{t}$ represents errors created by unusual changes that are not supported by the model.

There are two trend models for $g(t)$. These include a saturating growth model and a piecewise linear model. A saturating growth model typically handles non-linear prediction, which meets our requirement. In the proposed NRM, we therefore use the saturating growth model for predicting the spread of the virus. The saturating growth model is represented as follows:
\begin{equation}
g(t)=\frac{C}{1+\exp (-k(t-m))}
\end{equation}

where $C$ is the carrying capacity; $k$ is the growth rate and $m$ is the offset parameter. However, the growth rate is not constant, and therefore NRM incorporate trend changes in the growth model by defining change points where the growth rate can change. This is done by defining a vector of rate adjustments as follows [30] [31]:
\begin{equation}
\delta \in \mathbb{R}^{S}
\end{equation}

where $S$ represents change points at times and can be seen as $s_{j}$, $j$ = 1,…, $S$; $\delta_{j}$ is the change in rate that occurs at $s_{j}$ [31].

When the rate at time $t$ is equal to $k+a(t)^{T} \delta$. Then $k$ is adjusted, the offset parameter $m$ must also be adjusted to connect endpoints of segments. When there is a correct adjustment $\gamma_{j}$ at change point $j$, it can be computed as [31]:
\begin{equation}
\gamma_{j}=\left(s_{j}-m-\sum_{l<j} \gamma_{l}\right)\left(1-\frac{k+\sum_{l<j} \delta_{l}}{k+\sum_{l \leq j} \delta_{l}}\right)
\end{equation}

Finally, the model for logistic growth is given by the following equation:
\begin{equation}
g(t)=\frac{C(t)}{1+\exp \left(-\left(k+\mathbf{a}(t)^{\top} \boldsymbol{\delta}\right)\left(t-\left(m+\mathbf{a}(t)^{\top} \boldsymbol{\gamma}\right)\right)\right)}
\end{equation}

The proposed NRM was trained and tested in the same way as the DSPM, however, without using scalars for data input vectors.

\begin{table*}[]
\begin{center}
  \begin{tabular}{|l|l|l|l|l|l|}
\hline
Province/                   & Country/                   & Ground Truth & Baseline  & DSPM  & NRM  \\
State                   & Region                   & (Confirmed Cases) &  Prediction & Prediction & Prediction \\ \hline
                                 & Afghanistan                      & 16509                                              & 14              & 15537           & 15082           \\ \hline
                                 & Albania                          & 1164                                               & 203             & 1114            & 1124            \\ \hline
                                 & Algeria                          & 9626                                               & 132             & 9805            & 9822            \\ \hline
                                 & Andorra                          & 844                                                & 275             & 721             & 930              \\ \hline
                                 & Angola                           & 86                                                 & 115             & 90              & 81              \\ \hline
                                 & Antigua and Barbuda              & 26                                                 & 285              & 23              & 30              \\ \hline
                                 & Argentina                        & 18319                                              & 75              & 18290           & 16423           \\ \hline
                                 & Armenia                          & 10009                                              & 68              & 9969            & 9165            \\ \hline
Australian Capital Territory     & Australia                        & 107                                                & 315                & 101             & 131             \\ \hline
New South Wales                  & Australia                        & 3104                                               & 310             & 3077            & 3773            \\ \hline
Northern Territory               & Australia                        & 29                                                 & 301             & 28              & 35              \\ \hline
Queensland                       & Australia                        & 1059                                               & 310             & 1017             & 1290            \\ \hline
South Australia                  & Australia                        & 440                                                & 319             & 455             & 540             \\ \hline
Tasmania                         & Australia                        & 228                                                & 282             & 2228             & 229             \\ \hline
Victoria                         & Australia                        & 1670                                               & 261             & 1602            & 1883            \\ \hline
Western Australia                & Australia                        & 592                                                & 299                & 599            & 688             \\ \hline
                                 & Austria                          & 16759                                              & 295             & 16220           & 19713            \\ \hline
                                 & Azerbaijan                       & 5935                                               & 99              & 5565            & 5439            \\ \hline
                                 & Bahamas                          & 102                                                & 233             & 97              & 103              \\ \hline
                                 & Bahrain                          & 12311                                              & 77              & 12245           & 11732           \\ \hline
                                 & Bangladesh                       & 52445                                              & 1               & 52082           & 47162           \\ \hline
                                 & Barbados                         & 92                                                 & 274             & 90              & 94              \\ \hline
                                 & Belarus                          & 44255                                              & 14              & 44662           & 44491           \\ \hline
                                 & Belgium                          & 58615                                              & 246             & 56654          & 59266           \\ \hline
                                 & Benin                            & 244                                                & 24             & 217                & 240             \\ \hline
                                 & Bhutan                           & 47                                                 & 44             & 37              & 37              \\ \hline
                                 & Bolivia                          & 10991                                              & 12              & 13123           & 9453           \\ \hline
                                 & Bosnia and Herzegovina           & 2535                                               & 211             & 2538            & 2569            \\ \hline
                                 & Brazil                           & 555383                                             & 14              & 578432          & 509319          \\ \hline
                                 & Brunei                           & 141                                                & 66                 & 138             & 1623            \\ \hline
                                 & Bulgaria                         & 2538                                               & 181             & 2375            & 2635            \\ \hline
                                 & Burkina Faso                     & 881                                                & 227             & 786                & 877             \\ \hline
                                 & Cabo Verde                       & 466                                                & 13              & 409            & 463             \\ \hline
                                 & Cambodia                         & 125                                                & 64              & 125            & 151            \\ \hline
                                 & Cameroon                         & 6585                                               & 86             & 6758          & 6135            \\ \hline
Alberta                          & Canada                           & 7057                                               & 204             & 6736            & 7169            \\ \hline
British Columbia                 & Canada                           & 2601                                               & 259             & 2474            & 2622           \\ \hline
Grand Princess                   & Canada                           & 13                                                 & 66                 & 12             & 17              \\ \hline
Manitoba                         & Canada                           & 297                                                & 286             & 306             & 348             \\ \hline
New Brunswick                    & Canada                           & 133                                                & 276             & 124             & 149             \\ \hline
Newfoundland and Labrador        & Canada                           & 261                                                & 308             & 247             & 317             \\ \hline
Nova Scotia                      & Canada                           & 1057                                               & 260             & 1054            & 1066           \\ \hline
Ontario                          & Canada                           & 30259                                              & 171             & 28005          & 30191           \\ \hline
Prince Edward Island             & Canada                           & 27                                                 & 308             & 25              & 32              \\ \hline
Quebec                           & Canada                           & 51593                                              & 169             & 49344           & 52995           \\ \hline
Saskatchewan                     & Canada                           & 646                                                & 213             & 614             & 699             \\ \hline
                                 & Central African Republic         & 1069                                               & 1               & 1087            & 870             \\ \hline
                                 & Chad                             & 803                                                & 11              & 857            & 851             \\ \hline
                                 & Chile                            & 108686                                             & 49              & 109760           & 98245          \\ \hline
Anhui                            & China                            & 991                                                & 131             & 908             & 990             \\ \hline
Beijing                          & China                            & 593                                                & 166             & 575            & 593            \\ \hline
Chongqing                        & China                            & 579                                                & 130             & 567             & 579             \\ \hline
Fujian                           & China                            & 358                                                & 159             & 343             & 358             \\ \hline
Gansu                            & China                            & 139                                                & 168             & 136             & 140             \\ \hline
Guangdong                        & China                            & 1597                                               & 157             & 1519            & 1594           \\ \hline
Guangxi                          & China                            & 254                                                & 130             & 241             & 254             \\ \hline
Guizhou                          & China                            & 147                                                & 130            & 138            & 147             \\ \hline
Hainan                           & China                            & 169                                                & 129             & 158             & 168             \\ \hline
Hebei                            & China                            & 328                                                & 136            & 299            & 327             \\ \hline
Heilongjiang                     & China                            & 945                                                & 233            & 892             & 1072           \\ \hline
Henan                            & China                            & 1276                                               & 130             & 1229            & 1274            \\ \hline
Hong Kong                        & China                            & 1093                                               & 306            & 1050          & 1231             \\ \hline
Hubei                            & China                            & 68135                                              & 130           & 62677           & 67824          \\ \hline
Hunan                            & China                            & 1019                                               & 131             & 1001            & 1017            \\ \hline
Inner Mongolia                   & China                            & 235                                                & 213             & 235           & 244            \\ \hline
Jiangsu                          & China                            & 653                                                & 140             & 607             & 652             \\ \hline
Jiangxi                          & China                            & 937                                                & 132            & 897             & 936             \\ \hline
Jilin                            & China                            & 155                                                & 118            & 147            & 163             \\ \hline
Liaoning                         & China                            & 149                                                & 156           & 133            & 149            \\ \hline
Macau                            & China                            & 45                                                 & 281             & 41              & 51             \\ \hline
Ningxia                          & China                            & 75                                                 & 131            & 73             & 75             \\ \hline
 \hline
\end{tabular}
\vspace{2mm}
\caption{Prediction of Confirmed cases by our proposed models and the baseline approach.}
\label{table1}
 
\end{center}
\end{table*}

\begin{table*}[]
\begin{center}
\begin{tabular}{|l|l|l|l|l|l|}
\hline
Province/                   & Country/                   & Ground Truth & Baseline  & DSPM  & NRM  \\
State                   & Region                   & (Confirmed Cases) &  Prediction & Prediction & Prediction \\ \hline
          Qinghai                          & China                            & 18                                                 & 132             & 18             & 19             \\ \hline
Shaanxi                          & China                            & 309                                                & 159             & 291             & 309             \\ \hline
Shandong                         & China                            & 792                                                & 136             & 794            & 791             \\ \hline
Shanghai                         & China                            & 673                                                & 223             & 622             & 699             \\ \hline
Shanxi                           & China                            & 198                                                & 205              & 190             & 213            \\ \hline
Sichuan                          & China                            & 577                                                & 133             & 521            & 567                \\ \hline
Tianjin                          & China                            & 192                                                & 171             & 172          & 192             \\ \hline
Tibet                            & China                            & 1                                                  & 135             & 1               & 1             \\ \hline
Xinjiang                         & China                            & 76                                                 & 130             & 73              & 76             \\ \hline
Yunnan                           & China                            & 185                                                & 145             & 183             & 185            \\ \hline
Zhejiang                         & China                            & 1268                                               & 137             & 1209            & 1265          \\ \hline
                                 & Colombia                         & 30593                                              & 65              & 31723           & 28414           \\ \hline
                                 & Congo (Brazzaville)              & 611                                                & 112             & 570            & 608            \\ \hline
                                 & Congo (Kinshasa)                 & 3326                                               & 52              & 3689            & 3051            \\ \hline
                                 & Costa Rica                       & 1105                                               & 206             & 1128           & 1058               \\ \hline
                                 & Cote d'Ivoire                    & 3024                                               & 131            & 2971           & 2922           \\ \hline
                                 & Croatia                          & 2246                                               & 286             & 2123           & 2273             \\ \hline
                                 & Diamond Princess                 & 712                                                & 131             & 6781             & 710             \\ \hline
                                 & Cuba                             & 2092                                               & 211             & 2019            & 2068            \\ \hline
                                 & Cyprus                           & 952                                                & 268             & 880              & 953            \\ \hline
                                 & Czechia                          & 9364                                               & 256            & 8484             & 9375            \\ \hline
Faroe Islands                    & Denmark                          & 187                                                & 66                 & 165             & 233             \\ \hline
Greenland                        & Denmark                          & 13                                                 & 271             & 12             & 14             \\ \hline
                                 & Denmark                          & 11734                                              & 240             & 11976           & 11882           \\ \hline
                                 & Djibouti                         & 3779                                               & 14              & 3345           & 2965            \\ \hline
                                 & Dominican Republic               & 17752                                              & 126             & 17700         & 17759          \\ \hline
                                 & Ecuador                          & 40414                                              & 102            & 38943          & 42524          \\ \hline
                                 & Egypt                            & 27536                                              & 63              & 26582           & 23998              \\ \hline
                                 & El Salvador                      & 2653                                               & 15              & 2778            & 2582            \\ \hline
                                 & Equatorial Guinea                & 1306                                               & 12              & 1303            & 1282           \\ \hline
                                 & Eritrea                          & 39                                                 & 308             & 39              & 47              \\ \hline
                                 & Estonia                          & 1870                                               & 281             & 1716            & 1871            \\ \hline
                                 & Eswatini                         & 294                                                & 45              & 275           & 301             \\ \hline
                                 & Ethiopia                         & 1344                                               & 35                 & 1702            & 917             \\ \hline
                                 & Fiji                             & 18                                                 & 303             & 17              & 21              \\ \hline
                                 & Finland                          & 6887                                               & 220             & 6355            & 7006            \\ \hline
French Guiana                    & France                           & 517                                                & 75              & 541             & 439            \\ \hline
French Polynesia                 & France                           & 60                                                 & 310             & 57              & 72              \\ \hline
Guadeloupe                       & France                           & 162                                                & 302             & 159           & 191             \\ \hline
Mayotte                          & France                           & 1986                                               & 76              & 1885            & 1941           \\ \hline
New Caledonia                    & France                           & 20                                                 & 59              & 19               & 22               \\ \hline
Reunion                          & France                           & 477                                                & 281            & 439           & 542             \\ \hline
Saint Barthelemy                 & France                           & 6                                                  & 66                 & 6              & 6               \\ \hline
St Martin                        & France                           & 41                                                 & 281             & 40              & 47              \\ \hline
Martinique                       & France                           & 200                                                & 283            & 181            & 231             \\ \hline
                                 & France                           & 184980                                             & 265            & 177107           & 186533             \\ \hline
                                 & Gabon                            & 2803                                               & 2               & 2998           & 2813           \\ \hline
                                 & Gambia                           & 25                                                 & 179             & 24              & 28              \\ \hline
                                 & Georgia                          & 796                                                & 202             & 749            & 788             \\ \hline
                                 & Germany                          & 183879                                             & 276            & 157952           & 184833          \\ \hline
                                 & Ghana                            & 8297                                               & 16              & 7860            & 8553            \\ \hline
                                 & Greece                           & 2937                                               & 281             & 2811            & 2964           \\ \hline
                                 & Guatemala                        & 5586                                               & 12              & 6306            & 4877           \\ \hline
                                 & Guinea                           & 3886                                               & 16              & 3719           & 3891            \\ \hline
                                 & Guyana                           & 153                                                & 163             & 150             & 154           \\ \hline
                                 & Haiti                            & 2226                                               & 13             & 2758           & 1493           \\ \hline
                                 & Holy See                         & 12                                                 & 272             & 11             & 12             \\ \hline
                                 & Honduras                         & 5527                                               & 48              & 5728            & 5283           \\ \hline
                                 & Hungary                          & 3921                                               & 210             & 3811           & 3972            \\ \hline
                                 & Iceland                          & 1806                                               & 326             & 1743           & 2202           \\ \hline
                                 & India                            & 207191                                             & 14              & 219792         & 191044          \\ \hline
                                 & Indonesia                        & 27549                                              & 116             & 27994         & 27137.83           \\ \hline
                                 & Iran                             & 157562                                             & 193             & 148252           & 154378         \\ \hline
                                 & Iraq                             & 7387                                               & 91              & 7028            & 6076           \\ \hline
                                 & Ireland                          & 25066                                              & 233             & 23658           & 25437           \\ \hline
                                 & Israel                           & 17285                                              & 282             & 16998          & 17042          \\ \hline
                                 & Italy                            & 233515                                             & 264             & 227832           & 235225             \\ \hline
                                 & Jamaica                          & 590                                                & 191             & 550             & 587             \\ \hline
                                 & Japan                            & 16837                                              & 239             & 15845           & 16954          \\ \hline
                                 & Jordan                           & 755                                                & 193             & 682           & 780            \\ \hline
                                 & Kazakhstan                       & 11571                                              & 87              & 11734           & 10796           \\ \hline
                              
\end{tabular}
\vspace{2mm}
\caption{Prediction of Confirmed cases by our proposed models and the baseline approach.}
\end{center}
\end{table*}

\begin{table*}[]
\begin{center}

\begin{tabular}{|l|l|l|l|l|l|}
\hline
Province/                   & Country/                   & Ground Truth & Baseline  & DSPM  & NRM  \\
State                   & Region                   & (Confirmed Cases) &  Prediction & Prediction & Prediction \\ \hline
                                       & Kenya                            & 2093                                               & 65               & 2170            & 1787.          \\ \hline
                                    & Korea, South                     & 11590                                              & 62              & 10794           & 11612           \\ \hline
                                 & Kuwait                           & 28649                                              & 14              & 30006           & 28635           \\ \hline
                                 & Kyrgyzstan                       & 1845                                               & 126            & 1953           & 1749           \\ \hline
                                 & Latvia                           & 1071                                               & 254             & 1031            & 1092            \\ \hline
                                 & Lebanon                          & 1242                                               & 175            & 1195           & 1247           \\ \hline
                                 & Liberia                          & 311                                                & 16              & 297             & 298            \\ \hline
                                 & Liechtenstein                    & 82                                                 & 66                 & 80             & 100             \\ \hline
                                 & Lithuania                        & 1682                                               & 256             & 1635           & 1695           \\ \hline
                                 & Luxembourg                       & 4020                                               & 299             & 3676           & 4777           \\ \hline
                                 & Madagascar                       & 845                                                & 54            & 890           & 713            \\ \hline
                                 & Malaysia                         & 7877                                               & 243             & 6644           & 7860           \\ \hline
                                 & Maldives                         & 1841                                               & 10.             & 1832          & 1771           \\ \hline
                                 & Malta                            & 620                                                & 239             & 587            & 638             \\ \hline
                                 & Mauritania                       & 668                                                & 1               & 616            & 373            \\ \hline
                                 & Mauritius                        & 335                                                & 303             & 337             & 405             \\ \hline
                                 & Mexico                           & 97326                                              & 15              & 102505           & 92686           \\ \hline
                                 & Moldova                          & 8548                                               & 137             & 7946            & 8460            \\ \hline
                                 & Monaco                           & 99                                                 & 308            & 94              & 118            \\ \hline
                                 & Mongolia                         & 185                                                & 62              & 188             & 184             \\ \hline
                                 & Montenegro                       & 324                                                & 304             & 314            & 391             \\ \hline
                                 & Morocco                          & 7866                                               & 181              & 7422            & 8233           \\ \hline
                                 & Namibia                          & 25                                                 & 205             & 22              & 23             \\ \hline
                                 & Nepal                            & 2099                                               & 12             & 2389            & 1155            \\ \hline
Aruba                            & Netherlands                      & 101                                                & 309             & 96              & 123             \\ \hline
Curacao                          & Netherlands                      & 20                                                 & 256             & 18             & 18              \\ \hline
Sint Maarten                     & Netherlands                      & 77                                                 & 303            & 75             & 94              \\ \hline
                                 & Netherlands                      & 46647                                              & 250             & 45743          & 46858          \\ \hline
                                 & New Zealand                      & 1504                                               & 302            & 1532           & 1816            \\ \hline
                                 & Nicaragua                        & 1118                                               & 2               & 1161           & 556                \\ \hline
                                 & Niger                            & 960                                                & 240           & 906             & 1005            \\ \hline
                                 & Nigeria                          & 10819                                              & 13.72              & 10869           & 10386           \\ \hline
                                 & North Macedonia                  & 2391                                               & 179            & 2265           & 2233           \\ \hline
                                 & Norway                           & 8455                                               & 299             & 8135          & 8488            \\ \hline
                                 & Oman                             & 12799                                              & 15              & 15722          & 10678          \\ \hline
                                 & Pakistan                         & 76398                                              & 65              & 82776          & 71462           \\ \hline
                                 & Panama                           & 14095                                              & 142            & 12861          & 13208          \\ \hline
                                 & Papua New Guinea                 & 8                                                  & 254             & 8             & 9               \\ \hline
                                 & Paraguay                         & 1013                                               & 93              & 986             & 1053            \\ \hline
                                 & Peru                             & 170039                                             & 14              & 172132          & 162847           \\ \hline
                                 & Philippines                      & 18997                                              & 139            & 17515              & 17397            \\ \hline
                                 & Poland                           & 24395                                              & 160             & 23409           & 24561           \\ \hline
                                 & Portugal                         & 32895                                              & 225             & 30793          & 32822           \\ \hline
                                 & Qatar                            & 60259                                              & 15             & 60627           & 58434           \\ \hline
                                 & Romania                          & 19517                                              & 192             & 18790          & 19817           \\ \hline
                                 & Russia                           & 423186                                             & 14             & 413939         & 429051          \\ \hline
                                 & Rwanda                           & 384                                                & 181            & 386           & 372           \\ \hline
                                 & Saint Lucia                      & 18                                                 & 285            & 17              & 21              \\ \hline
                                 & Saint Vincent and the Grenadines & 26                                                 & 169             & 26             & 23              \\ \hline
                                 & San Marino                       & 672                                                & 271            & 640              & 683            \\ \hline
                                 & Saudi Arabia                     & 89011                                              & 16              & 83758           & 90707           \\ \hline
                                 & Senegal                          & 3836                                               & 65             & 3992           & 3831          \\ \hline
                                 & Serbia                           & 11454                                              & 140             & 10816          & 11629           \\ \hline
                                 & Seychelles                       & 11                                                 & 66                 & 10             & 13             \\ \hline
                                 & Singapore                        & 35836                                              & 17.             & 37060         & 36446          \\ \hline
                                 & Slovakia                         & 1522                                               & 264             & 1434             & 1535            \\ \hline
                                 & Slovenia                         & 1475                                               & 315            & 1418               & 1475          \\ \hline
                                 & Somalia                          & 2089                                               & 2               & 1952           & 2068             \\ \hline
                                 & South Africa                     & 35812                                              & 51              & 39859         & 32038          \\ \hline
                                 & Spain                            & 239932                                             & 276             & 240586           & 241540           \\ \hline
                                 & Sri Lanka                        & 1683                                               & 117             & 1681            & 1535            \\ \hline
                                 & Sudan                            & 5310                                               & 8               & 5750           & 5204            \\ \hline
                                 & Suriname                         & 54                                                 & 12              & 42            & 17             \\ \hline
                                 & Sweden                           & 38589                                              & 165            & 36476           & 38396           \\ \hline
                                 & Switzerland                      & 30874                                              & 306             & 28365          & 36785          \\ \hline
                                 & Taiwan*                          & 443                                                & 297            & 426           & 498           \\ \hline
                                 & Tanzania                         & 509                                                & 18             & 496           & 616           \\ \hline
                                 & Thailand                         & 3083                                               & 304             & 3134          & 3677          \\ \hline
                                 & Togo                             & 445                                                & 84              & 442            & 472             \\ \hline
                                 & Trinidad and Tobago              & 117                                                & 308             & 113            & 143            \\ \hline
 \hline
\end{tabular}
\vspace{2mm}
\caption{Prediction of Confirmed cases by our proposed models and the baseline approach.}
\end{center}
\end{table*}

\begin{table*}[]
\begin{center}
\begin{tabular}{|l|l|l|l|l|l|}
\hline
Province/                   & Country/                   & Ground Truth & Baseline  & DSPM  & NRM  \\
State                   & Region                   & (Confirmed Cases) &  Prediction & Prediction & Prediction \\ \hline
                                    
                                 & Turkey                           & 165555                                             & 211             & 157765        & 167550           \\ \hline
                                 & Uganda                           & 489                                                & 57              & 600            & 353             \\ \hline
                                 & Ukraine                          & 24895                                              & 123             & 25991         & 24664           \\ \hline
                                 & United Arab Emirates             & 35788                                              & 17              & 35526         & 36112         \\ \hline
Bermuda                          & United Kingdom                   & 141                                                & 224            & 135            & 140             \\ \hline
Cayman Islands                   & United Kingdom                   & 151                                                & 141             & 152             & 145           \\ \hline
Channel Islands                  & United Kingdom                   & 560                                                & 292            & 507           & 671            \\ \hline
Gibraltar                        & United Kingdom                   & 172                                                & 260            & 160             & 183           \\ \hline
Isle of Man                      & United Kingdom                   & 336                                                & 290         & 318          & 409             \\ \hline
Montserrat                       & United Kingdom                   & 11                                                 & 304              & 10             & 13          \\ \hline
                                 & United Kingdom                   & 277985                                             & 190            & 252623          & 284441             \\ \hline
                                 & Uruguay                          & 826                                                & 248           & 801            & 828            \\ \hline
                                 & US                               & 1831821                                            & 178             & 1787290            & 1835810            \\ \hline
                                 & Uzbekistan                       & 3760                                               & 157          & 3761            & 3681          \\ \hline
                                 & Venezuela                        & 1819                                               & 57              & 1743          & 1457           \\ \hline
                                 & Vietnam                          & 328                                                & 251             & 323             & 338             \\ \hline
                                 & Zambia                           & 1089                                               & 33              & 1174          & 1194           \\ \hline
                                 & Zimbabwe                         & 206                                                & 47            & 317             & 109             \\ \hline
Diamond Princess                 & Canada                           & 1                                                 &       -        & 1             & 1               \\ \hline
                                 & Dominica                         & 18                                                 & 283            & 15             & 19              \\ \hline
                                 & Grenada                          & 23                                                 & 261             & 22              & 24             \\ \hline
                                 & Mozambique                       & 307                                                & 68              & 261             & 2556              \\ \hline
                                 & Syria                            & 123                                                & 104            & 131             & 106            \\ \hline
                                 & Timor-Leste                      & 24                                                 & 19                 & 21              & 29              \\ \hline
                                 & Belize                           & 18                                                 & 284             & 17            & 21              \\ \hline
                                 & Laos                             & 19                                                 & 305             & 18             & 23             \\ \hline
                                 & Libya                            & 182                                                & 92             & 168             & 110          \\ \hline
                                 & West Bank and Gaza               & 451                                                & 237              & 443           & 447             \\ \hline
                                 & Guinea-Bissau                    & 1339                                               & 10               & 1308           & 1441            \\ \hline
                                 & Mali                             & 1351                                               & 16             & 1308          & 1286           \\ \hline
                                 & Saint Kitts and Nevis            & 15                                                 & 307            & 14           & 18             \\ \hline
Northwest Territories            & Canada                           & 5                                                  & 303            & 5              & 6               \\ \hline
Yukon                            & Canada                           & 11                                                 & 305            & 11              & 13              \\ \hline
                                 & Kosovo                           & 1064                                               & 206             & 1033           & 1093           \\ \hline
                                 & Burma                            & 232                                                & 188             & 220         & 223          \\ \hline
Anguilla                         & United Kingdom                   & 3                                                  & 307           & 3               & 3           \\ \hline
British Virgin Islands           & United Kingdom                   & 8                                                  & 226             & 7              & 8               \\ \hline
Turks and Caicos Islands         & United Kingdom                   & 12                                                 & 304             & 11              & 14             \\ \hline
                                 & MS Zaandam                       & 9                                                  & 307             & 9               & 10            \\ \hline
                                 & Botswana                         & 40                                                 & 159             & 36              & 36             \\ \hline
                                 & Burundi                          & 63                                                 & 55              & 62             & 54              \\ \hline
                                 & Sierra Leone                     & 896                                                & 1             & 851            & 914            \\ \hline
Bonaire, Sint Eustatius and Saba & Netherlands                      & 7                                                  & 234            & 7               & 6             \\ \hline
                                 & Malawi                           & 358                                                & 13             & 625          & 180            \\ \hline
Falkland Islands (Malvinas)      & United Kingdom                   & 13                                                 & 132          & 13           & 15               \\ \hline
Saint Pierre and Miquelon        & France                           & 1                                                  & 309             & 1              & 1              \\ \hline
                                 & South Sudan                      & 994                                                & 5               & 1450           & 928           \\ \hline
                                 & Western Sahara                   & 9                                                  & 201             & 8              & 9             \\ \hline
                                 & Sao Tome and Principe            & 484                                                & 11              & 532             & 464             \\ \hline
                                 & Yemen                            & 399                                                & 3               & 0.16               & 347           \\ \hline
                                 & Comoros                          & 132                                                & 273               & 100               & 83              \\ \hline
                                 & Tajikistan                       & 4100                                               & 3                & 29             & 4103            \\ \hline
                                 & Lesotho                          & 2                                                  & 3               & 0                  & 2               \\ \hline
\end{tabular}
\vspace{2mm}
\caption{Prediction of Confirmed cases by our proposed models and the baseline approach.}
\label{table4}
\end{center}
\end{table*}
\section{Experimental Results}
We extensively evaluated the performance of the proposed models on the publicly available novel coronavirus (COVID-19) dataset. In this section, we first provide the details of the dataset and then present our experimental results.

\subsection{Novel Coronavirus 2019 Dataset}
We used publicly available novel Coronavirus 2019 dataset \cite{Khoong2020}, \cite{JHU2019}, which is available via Kaggle and Github, respectively. The dataset contains globally reported confirmed COVID-19 cases in the following format:
 
\textbf{ObservationDate} - Date of the observation in MM/DD/YYYY

\textbf{Province/State} - Province or state of the observation 

\textbf{Country/Region} - Country of observation

\textbf{Last Update} - Time in UTC at which the row is updated for the given province or country. 

\textbf{Confirmed} - Cumulative number of confirmed cases till that date

\textbf{Deaths} - Cumulative number of of deaths till that date

\textbf{Recovered} - Cumulative number of recovered cases till that date

In the dataset, there are 133 dates that are represented as time series points, and each time series point includes the number of confirmed COVID-19 cases on that date. There are 266 countries that are split up into provinces that have data for those 133 dates. There is also other data that includes recovery cases, and death cases that follow the same format as the confirmed cases. Our proposed models have been evaluated on 6.4 million confirmed COVID-19 cases, which have been reported from 22$^{nd}$ January to 5$^{th}$ May 2020. 

\subsection{Data Preprocessing}
The data fed to each model is divided into country and state/province level and stored in objects to allow easy access to country predictions and error rates. Some of the predictions are in decimal value. All these prediction values are rounded to the nearest whole number to represent the actual number of infected people.

\subsection{Metric for Evaluation}
Prediction values are compared to real cases by using Mean Absolute Error (MAE), which is a loss function mostly used for regression models. MAE is a metric that is used to compare both predicted value and the actual value. MAE is measured for each prediction, before the prediction values are rounded for computing an accurate error rate.

\subsection{Prediction Results}
In this section, we present the prediction results for the proposed models and comparison with the baseline model.

\subsection{Baseline Method and Results}
We use the popular Support Vector Machine (SVM) as our baseline method (called Model1 in our experiments) to predict and analyze the spread of coronavirus in different locations for a variety of reasons. One of the main reasons for choosing SVM was its ease of implementation. Using different Python algorithms allows for easy splitting of the data into training dataset and a test dataset, as well as the actual modelling of the dataset. Next, the model is good for showing and modelling linear and nonlinear (exponential) regression [9], meaning that it is able to model output variables that are real and/or continuous values, for example such as predicting the average age of a person [9], or in the case of this paper, predicting the spread of a coronavirus in a certain location. Lastly it is usually efficient, as it uses a subset of the data given as training data from the decision function, meaning that it is quick and memory efficient on smaller data sets.

\begin{figure*}
  \includegraphics[width=1\textwidth]{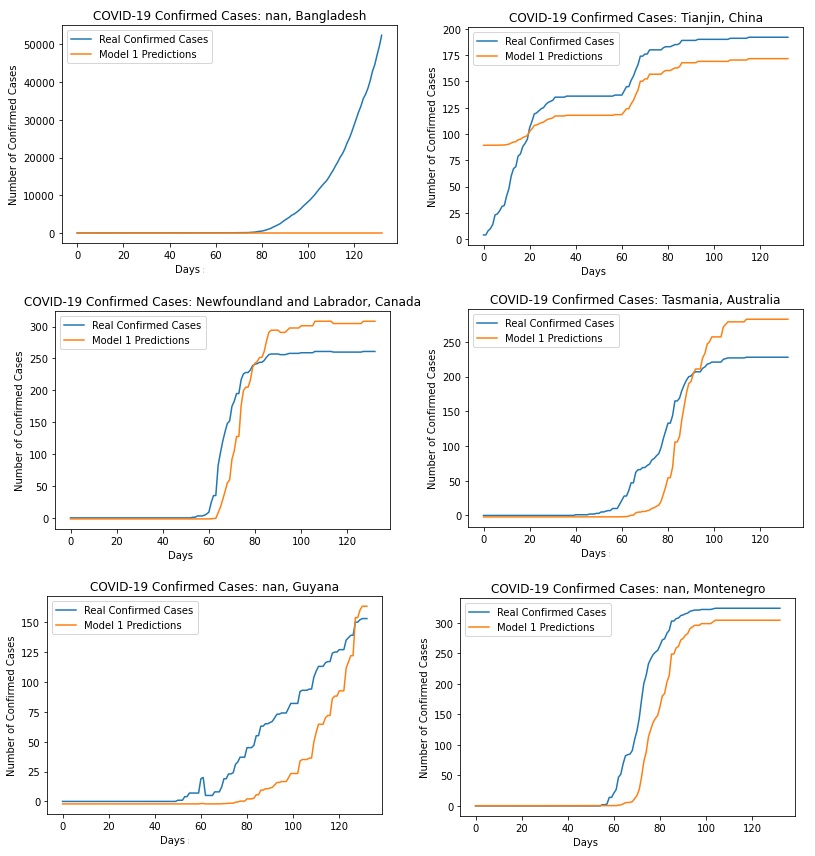}
\caption{Prediction results for the Baseline method (Model1). Countries/Regions have been randomly selected from the overall results to demonstrate the prediction performance of the baseline method. Additional prediction results are shown in Fig. \ref{fig:6}.}
\label{fig:3}       
\end{figure*}
\begin{figure*}
  \includegraphics[width=1\textwidth]{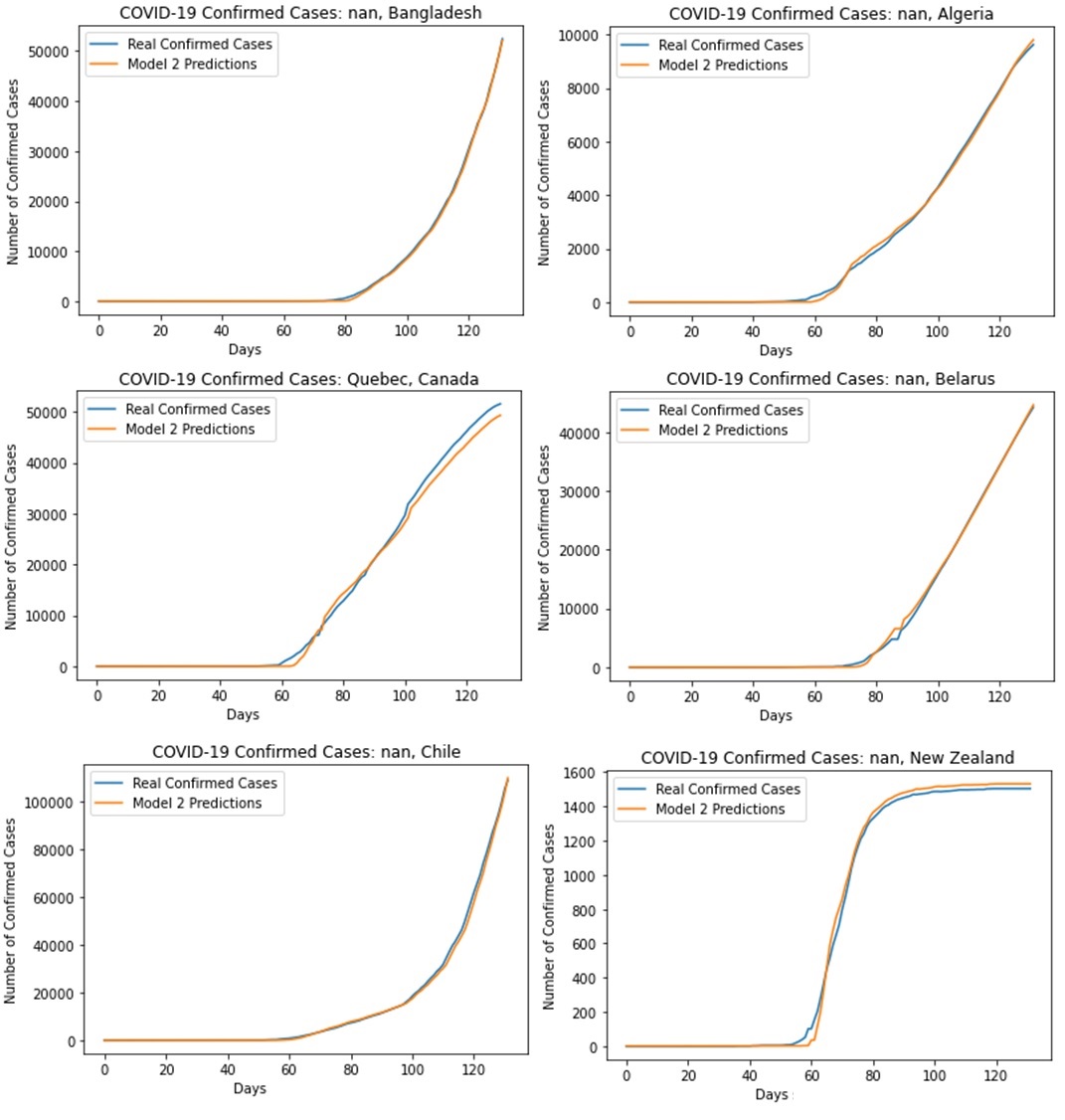}
\caption{Prediction results for the proposed Deep Sequential Prediction Model (Model2). Countries/Regions have been randomly selected from the overall results to demonstrate the prediction performance of the proposed DSPM. Additional prediction results are shown in Fig. \ref{fig:7}.}
\label{fig:4}       
\end{figure*}
\begin{figure*}
  \includegraphics[width=1\textwidth]{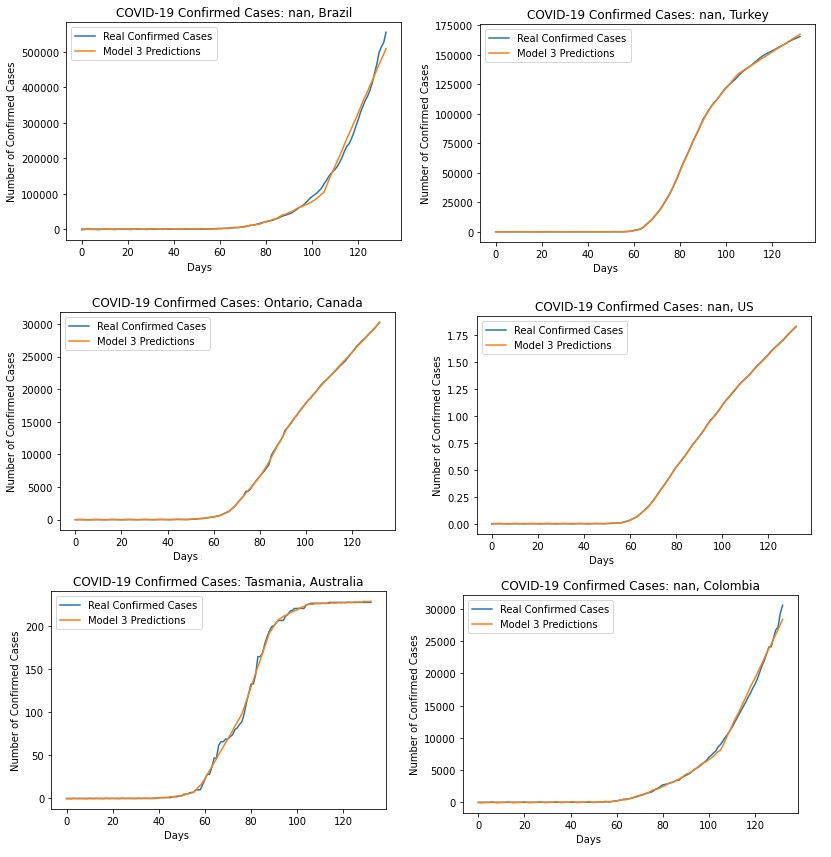}
\caption{Prediction results for the proposed Non-Parametric Regression Model (Model3). Countries/Regions have been randomly selected from the overall results to demonstrate the prediction performance of the proposed NRM. Additional prediction results are shown in Fig. \ref{fig:8}.}
\label{fig:5a}       
\end{figure*}
\begin{figure*}
  \includegraphics[width=1\textwidth]{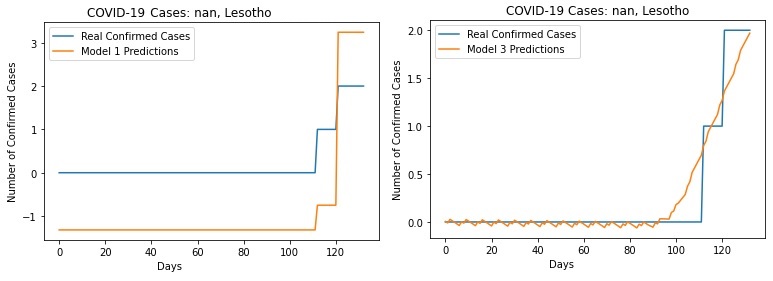}
\caption{Example of a country with low MAE and small number of COVID-19 cases.}
\label{fig:5}       
\end{figure*}
\begin{figure*}
  \includegraphics[width=0.9\textwidth]{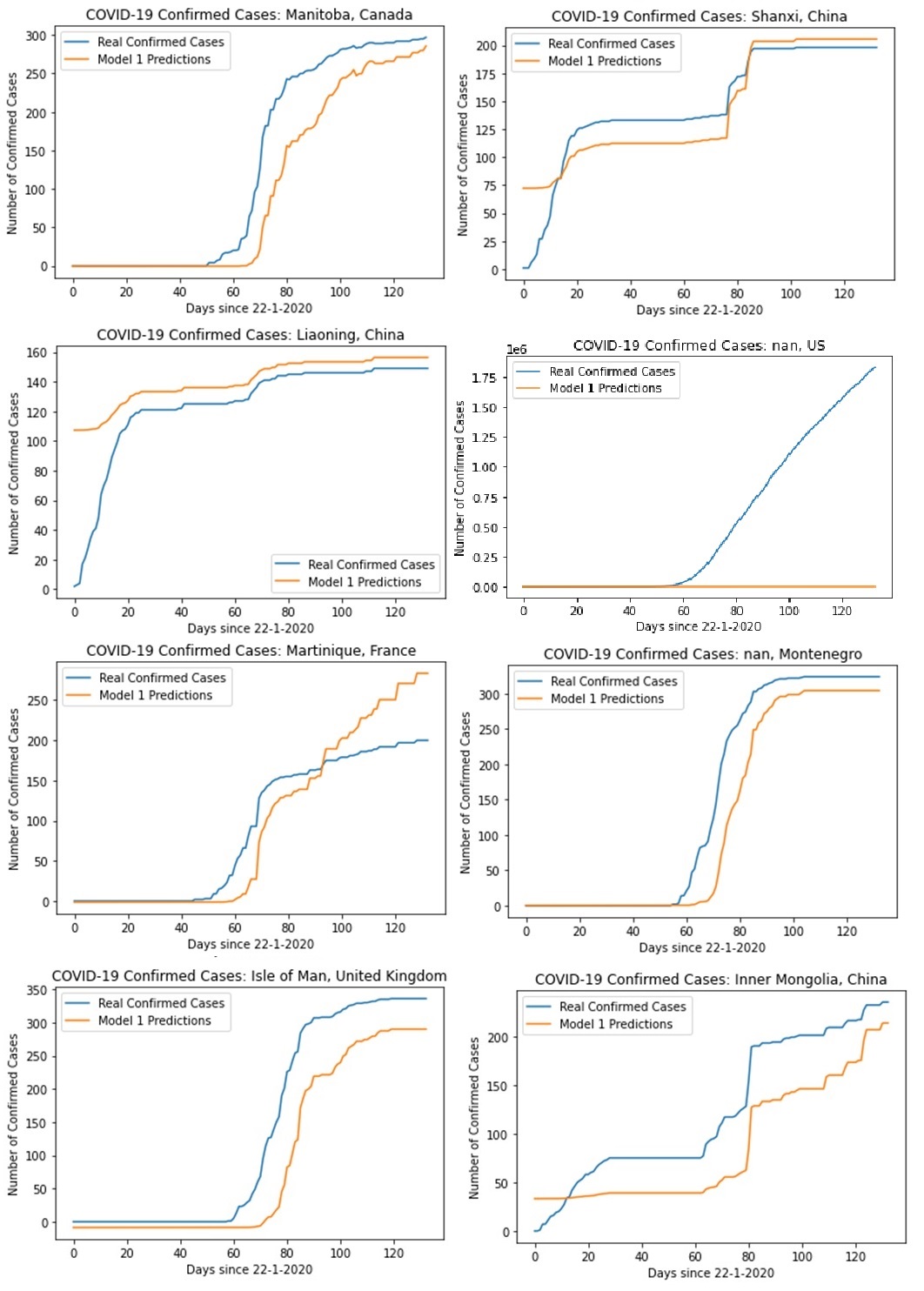}
\caption{Additional Prediction results for the baseline model (Model 1). Countries/Regions have been randomly selected from the overall results to demonstrate the prediction performance of the baseline.}
\label{fig:6}       
\end{figure*}

Table 1 to 4 (Column 4) report the SVM predictions formatted to be time-series data. Table 5 reports the average Mean Absolute Error (MAE) for the baseline model. As can be noted, the average MAE is really high compared to the total cases analyzed. Fig. 2 shows prediction results for the baseline model. Fig. 2 (first column and row) shown the country (Bangladesh) that has the highest MAE out of all countries that were analyzed by this model. It can be noted that this odel was not able to accurately predict COVID-19 cases for this country. A similar trend was observed for other countries that have a large number of confirmed corona virus cases. Fig.2 also shows countries with countries with better prediction results. Fig. 5 (left columns) shows the country that has the lowest MAE out of all countries that were analyzed. Low MAEs are usually found within countries that have the lowest number of confirmed cases. This can be seen in Fig. 5 for two different models, which have the lowest MAE for this country. It can be generalized that this model has a high failure rate when a country has large amount of cases to analyze. Table 5 reports the average MAE and error rate that can be expected as error estimate when the model predicts COVID-19 cases for a given country/region. Additional prediction results for this model have been provided in Fig. \ref{fig:6}.

\begin{table}
\centering
\begin{tabular}{@{}ccc@{}} 
\toprule
Model & Average MAE & Error Rate \\ \hline
Baseline (Model1) & 6508.22 & 27\%  \\ \hline
\textbf{Proposed DPSM} & 388.43  & 1.6\%\\ \hline 
\textbf{Proposed NRM} & 142.23  &  0.6\%   \\ 
 \bottomrule
\end{tabular}
\vspace{2mm}
\caption{MAE and error rates of our proposed models and the baseline approach.}
\label{table5}
\end{table}

\subsection{DSPM Results}
\begin{figure*}
  \includegraphics[width=0.9\textwidth]{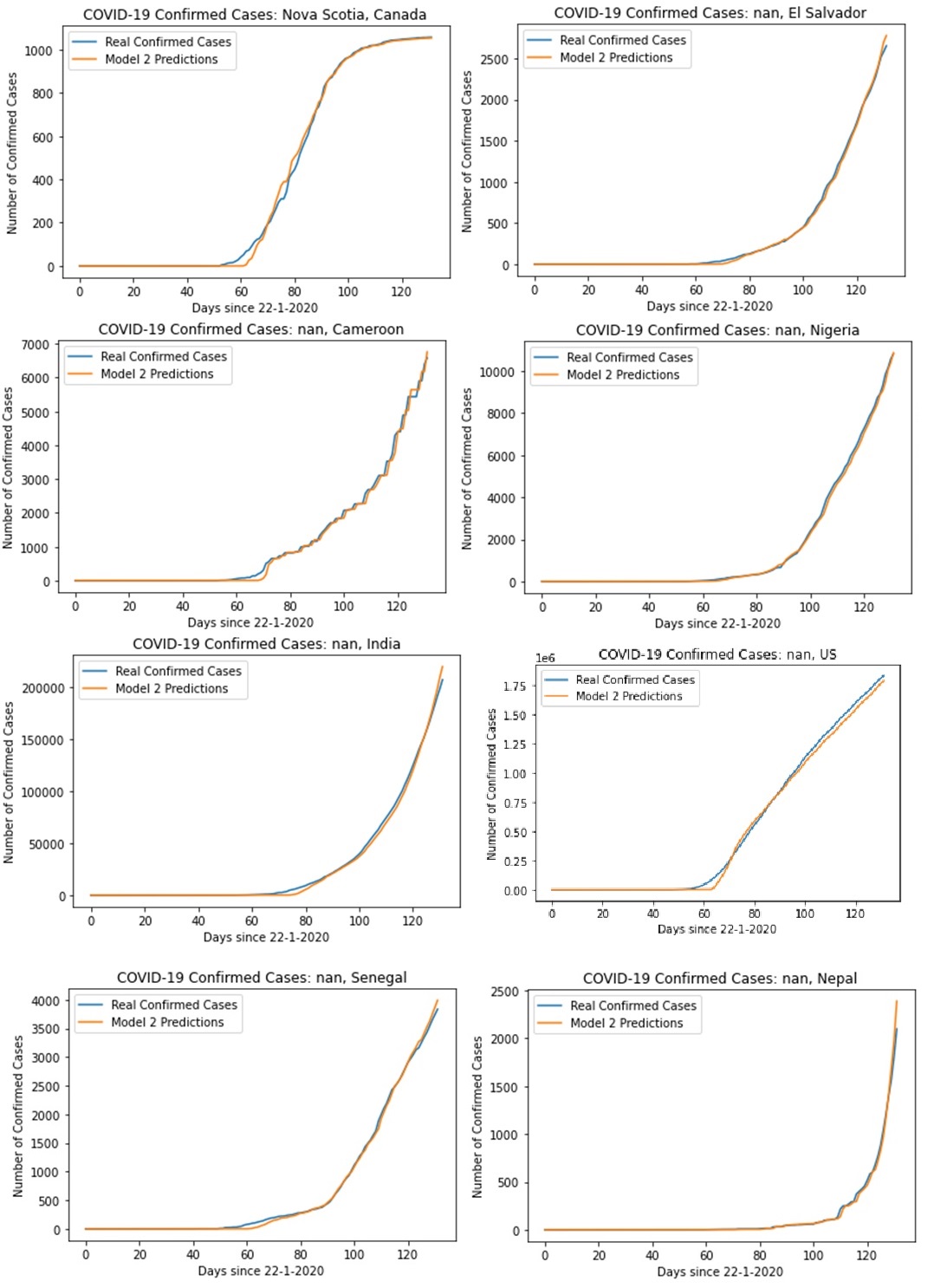}
\caption{Additional Prediction results for the proposed DPSM (Model 2). Countries/Regions have been randomly selected from the overall results to demonstrate the prediction performance of the proposed DPSM.}
\label{fig:7}       
\end{figure*}

Table 1 to 4 (Column 5) report the predictions results for the proposed DPSM that are formatted to be time-series data. Table 5 reports the prediction results for our proposed DSPM (called Model2 in our experiments). The average MAE for this model is 388.43, which is very low compared to the baseline model. The error rate for this model is 1.62\%. Fig. 3 shows the prediction results for the proposed DPSM. For this model, most countries and provinces with the lowest MAEs include countries and provinces that generally have lower cases of the virus (Fig. 3 and Table 1-4). Additional prediction results for this model have been provided in Fig. \ref{fig:7}.

\subsection{NRM Results}
\begin{figure*}
  \includegraphics[width=0.9\textwidth]{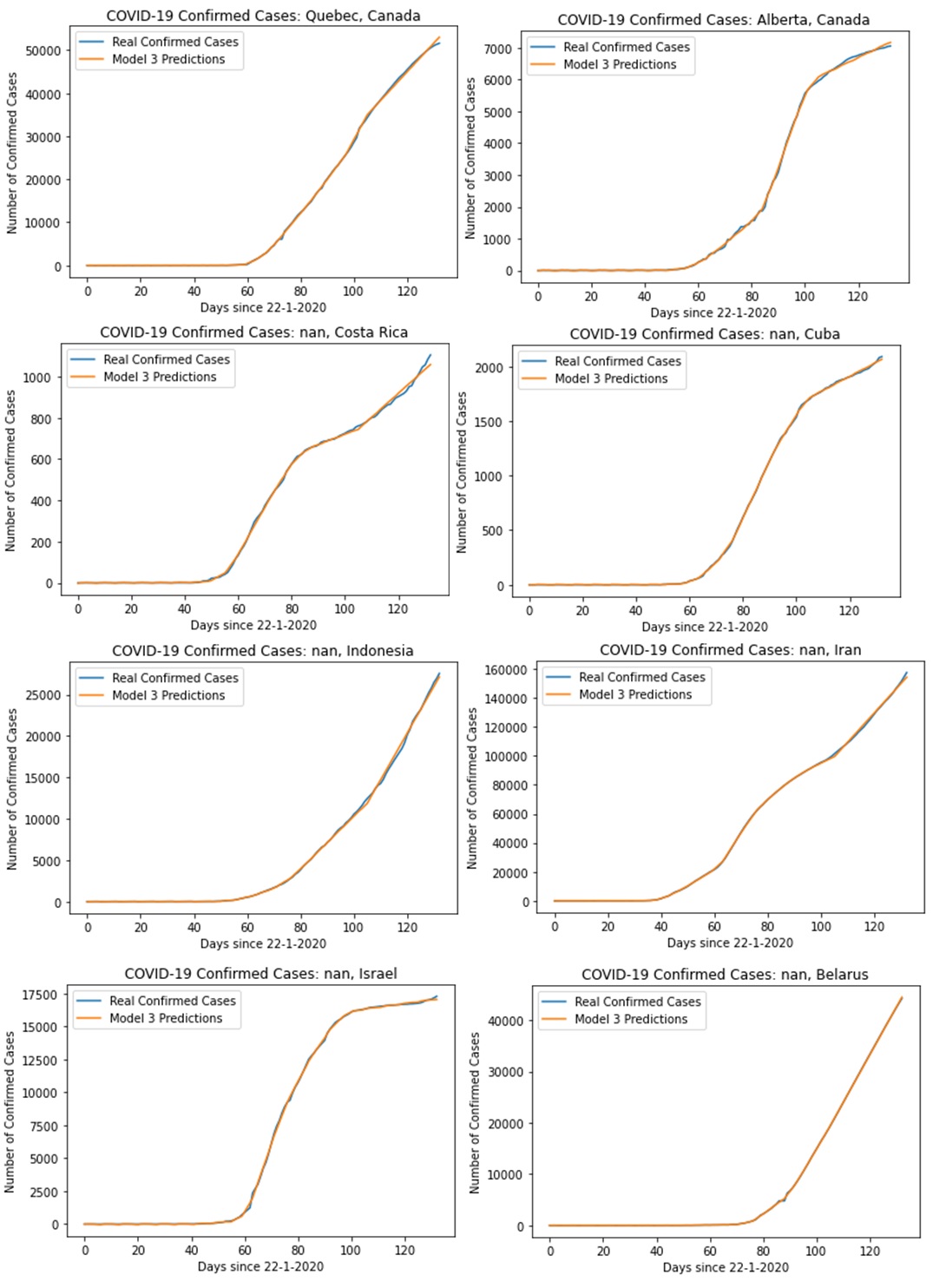}
\caption{Additional Prediction results for the proposed NRM (Model 3). Countries/Regions have been randomly selected from the overall results to demonstrate the prediction performance of the proposed NRM.}
\label{fig:8}       
\end{figure*}

Table 1 to 4 (Column 6) reports the prediction results for our proposed NRM (called Model 3 in our experiments). The average MAE for this model is 142.23 (Table 5), which is low compared to the baseline method and DPSM. The error rate for the proposed NRM is only 0.6\%. Fig. 4 shows the prediction results (randomly selected for demonstration) for this model. As can be noted this model achieves the best prediction results. The last row of Fig. 4 shows the countries and provinces that have the lowest error (calculated from MAE) in their continent. NRM outperforms the baseline model and DPSM. Additional prediction results for this model have been provided in Fig. \ref{fig:8}.

\section{Discussion and Analysis}
Table \ref{table5} reports average MAE for the baseline method and our proposed techniques. High MAEs generally do not always mean bad predictions. For instance in Fig. 4 (Brazil, first row and first column), there were 555383 confirmed cases analyzed in Brazil and having only a MAE error of 5472 basically means out of all the confirmed cases, 5472 individuals were predicted incorrectly. This means that there was only a 0.98\% error for the entire data for Brazil and overall this is a good prediction. High MAEs can be classified as a bad error rate for the model predictions when the error rate is over 10\% out of all confirmed cases for a country and province as seen in Fig. 2 (Bangladesh, first row and first column) for baseline methos (Model1). The MAE for this case is 522297.28 out of 1.83 million confirmed cases. The error in this case is 28.51\%. We observed that countries that have a small number of confirmed cases, generally have lower MAEs because there are not enough confirmed cases, thus models will have a limited range of cases that it can predict. This can be seen in Fig.\ref{fig:5} (for Lesotho), which shows different predictions for each model and both have low MAEs. Similar results are prevalent in other countries with small numbers of confirmed cases. 

Note that baseline model (Model1) has an error rate of 27\%, the proposed DPSM has an error rate of 1.62\% and the proposed NRM has an error rate of 0.6\%. Baseline model was not efficient enough compared to DPSM and NRM. In addition, proposed NRM performed better than the proposed DPSM, however, the difference in performance is not large. Both models can be used to model prediction for COVID-19 i.e., predict the number of people that can get infected by this disease. However, the models were only tested on the number of people being infected by Coronavirus and confirmed, it does not consider other factors such as recoveries, deaths, and restrictions being implemented that reduce the chances for a person contracting COVID-19. However, this does not limit the predictions that the models will make as they will follow trends that are continuously being updated within the provided COVID-19 dataset. 

\section{Conclusion and Future Work}
In this paper, deep/machine learning models have been developed with the purpose of accurately predicting the spread of COVID-19. These models include DPSM and NRM. The proposed models were trained and tested as predictive models for the spread of COVID-19. As can be noted, our proposed models were successful on predicting the spread of COVID-19 with low error rates. NRM was deemed the most accurate model to be used to predict the spread of the virus due to its low MAE and error rate (0.6\%), however the DPSM model was close to performing on the same level as NRM without any issues as it had lower overall error rates compared to cases per specific country and province. It can be concluded that the proposed DPSM and the NRM models have the potential to predict the spread of the virus in the future. However, the baseline model may have to be tweaked to fit time series data more efficiently and predict the spread with an overall lower MAE. COVID-19 was a virus that the world was poorly prepared for. The use of machine learning techniques as tools to predict the spread of the virus would allow for greater levels of preparedness through better resource management and distribution based on the prediction made by the models. These models can help prevent more waves of COVID-19 from occurring or even provide groundwork for the creation of similar predictive models for future strains of viruses.

In our future work, we intend to fuse DPSM and NRM features to refine the prediction of the proposed models. We would also train our model on additional data (as the publicly available dataset is being regularly updated) to further improve the prediction of the spread of COVID-19.

\section*{Acknowledgment}
This research is supported by Murdoch University Australia. 

%
%

\bibliographystyle{IEEEtran}    

\bibliography{main}   

%




\end{document}